\documentclass[10pt,twocolumn,letterpaper]{article}

\usepackage{iccv}
\usepackage{times}
\usepackage{epsfig}
\usepackage{graphicx}
\usepackage{amsmath}
\usepackage{amssymb}
\usepackage{multirow}
\usepackage{hhline}
\usepackage{caption}
\usepackage{subcaption}
\usepackage[demo]{rotating}
\usepackage{authblk}
\usepackage{float}

\newcommand{\eat}[1]{}

\usepackage{amsmath,amssymb,amsopn}
\usepackage{bm} % bold symbol
\usepackage{mathtools}

\usepackage[pagebackref=true,breaklinks=true,letterpaper=true,colorlinks,bookmarks=false]{hyperref}

\iccvfinalcopy % *** Uncomment this line for the final submission

\def\iccvPaperID{725} % *** Enter the ICCV Paper ID here

\ificcvfinal\fi
\begin{document}

\newcommand*\rot{\rotatebox{90}}

\newcommand{\BG}[1]{\textcolor{blue}{BG: #1}}
\newcommand{\PD}[1]{\textcolor{blue}{PD: #1}}
\newcommand{\YZ}[1]{\textcolor{blue}{YZ: #1}}

%%%%%%%%% TITLE
\title{Curriculum Domain Adaptation for Semantic Segmentation of Urban Scenes}

\iffalse
\author{Yang~Zhang\\
Center for Research in Computer Vision\\
University of Central Florida\\
{\tt\small yangzhang@knights.ucf.edu}
\and
Phillip~David\\
Computational and Information Sciences Directorate \\
U.S. Army Research Laboratory\\
{\tt\small philip.j.david4.civ@mail.mil}
\and
Boqing~Gong\\
Center for Research in Computer Vision\\
University of Central Florida\\
{\tt\small bgong@crcv.ucf.edu }
}
\fi

\author[1]{Yang~Zhang}
\author[2]{Philip~David}
\author[1]{Boqing~Gong}
\affil[1]{
Center for Research in Computer Vision, University of Central Florida
}
\affil[2]{
Computational and Information Sciences Directorate, U.S. Army Research Laboratory
{\tt\small yangzhang@knights.ucf.edu, philip.j.david4.civ@mail.mil, bgong@crcv.ucf.edu}
}

\maketitle
%\thispagestyle{empty}

%%%%%%%%% ABSTRACT
\begin{abstract}
    % !TEX root = main.tex

During the last half decade, convolutional neural networks (CNNs) have triumphed over semantic segmentation, which is a core task of various emerging industrial applications such as autonomous driving and medical imaging. However, to train CNNs requires a huge amount of data, which is difficult to collect and laborious to annotate. Recent advances in computer graphics make it possible to train CNN models on photo-realistic synthetic data with computer-generated annotations. Despite this, the domain mismatch between the real images and the synthetic data significantly decreases the models' performance. Hence we propose a curriculum-style learning approach to minimize the domain gap in semantic segmentation. The curriculum domain adaptation solves easy tasks first in order to infer some necessary properties about the target domain; in particular, the first task is to learn global label distributions over images and local distributions over landmark superpixels. These are easy to estimate because images of urban traffic scenes have strong idiosyncrasies (e.g., the size and spatial relations of buildings, streets, cars, etc.). We then train the segmentation network in such a way that the network predictions in the target domain follow those inferred properties.  In experiments, our method significantly outperforms the baselines as well as the only known existing approach to the same problem.

\vspace{-20pt}

%The former The latter prevents large area mismatch by anchoring the network prediction via few yet confidently labeled landmark superpixels; The global label distribution term fine-tuns object boundary for small objects by aligning the target prediction label distribution histogram with an estimated label distribution. 

\end{abstract}
\let\thefootnote\relax\footnote{\noindent For better reproducibility, the code is available at:\\ \url{https://github.com/YangZhang4065/AdaptationSeg}.}
%%%%%%%%% BODY TEXT
%!Tex root = main.tex

\section{Introduction}

This paper is concerned with domain adaptation for semantic image segmentation of urban scenes, i.e., assigning a category label to every pixel of an image or video frame~\cite{borenstein2002class}. Our interest in this problem is partially due to the exciting vision of autonomous driving, where understanding complex inner-city traffic scenes is an essential module and semantic segmentation is one of its key constituents~\cite{cordts_cityscapes_2016,geiger2013vision}. 

Machine learning methods for automatic semantic segmentation require massive amounts of high-quality annotated imagery in order to produce effective classifiers that generalize well to novel scenes. However, annotating training imagery for semantic segmentation is a very cumbersome task for humans. Cordts et al.\ report that the annotation and quality control take  more than 1.5 hours on a single image of the Cityscapes dataset~\cite{cordts_cityscapes_2016}. Besides, it is very difficult and time-consuming to collect imagery that depicts the large number of variabilities possible of urban scenes in different countries, seasons, and lighting conditions, etc. 

To overcome both shortcomings, simulated urban environments may be used to automatically generate large amounts of annotated training imagery. This, however, introduces a new problem, that of domain mismatch between the source (simulated) domain and the target (real) domain. Figure~\ref{fQualitative} illustrates some examples drawn from the synthetic SYNTHIA~\cite{ros_synthia_2016} dataset and the real Cityscapes~\cite{cordts_cityscapes_2016} dataset. It is readily apparent that there are significant visual differences between the two datasets. Domain adaptation techniques~\cite{ros_synthia_2016,shrivastava2016learning,hoffman_fcns_2016} may be used by machine learning methods to bridge this gap between the two domains.

In computer vision, learning domain-invariant features has been a prevalent and successful strategy to tackle the discrepancy between two domains, mainly for classification and regression problems~\cite{PanTKDE10Survey,PatelSPM15Visual}. The core idea is to infer a new feature space such that the marginal distributions of the source domain (S) and the target domain (T) are about the same, i.e., $P_S(Z)\approx P_T(Z)$. Furthermore, the prediction function $P(Y|Z)$ from that space  is assumed to be the same across the domains so that one can leverage the rich labeled data in the source domain to train classifiers that generalize well to the target. It is hard to verify the assumption, but the work along this line is rich and has led to impressive practical results regardless, such as the algorithms using linear transformation~\cite{GongCVPR12Geodesic,GopalanICCV11Domain,FernandoICCV13Unsupervised,SunAAAI16Return}, kernel methods~\cite{PanTNN11Domain,GongICML13Connecting,AljundiCVPR15Landmarks,KulisCVPR11What}, and the recent deep learning methods that directly extract domain-invariant features from raw input images~\cite{TzengX14Deep,LongICML15Learning,TzengICCV15Simultaneous,GaninX15Domainadversarial,GaninICML15Unsupervised}. 

In contrast to prior arts, the semantic segmentation we study in this paper is a  highly structured prediction problem, for which domain adaptation is only sparsely explored in the literature~\cite{yamada2014domain,hoffman_fcns_2016}. Under structured prediction, can we still achieve good domain adaptation results by  following the above principles? Our intuition and experimental studies (cf.\ Section~\ref{sExperiments}) tell us no. Learning a decision function for structured prediction is more involved than classification because it has to resolve the predictions in an exponentially large label space. As a result, the assumption that the source and target domains share the same prediction function becomes less likely to hold. Besides, some discriminative cues in the data would be suppressed if one matches the feature representations of the two domains without taking careful account of the structured labels. Finally, data instances are the proxy to measure the domain difference~\cite{GrettonBC09Covariate,GaninICML15Unsupervised,GaninX15Domainadversarial}. However, it is not immediately clear what comprises the instances in semantic segmentation~\cite{hoffman_fcns_2016}, especially given that the top-performing segmentation methods  are built upon deep neural networks~\cite{long_fully_2015,pathak2015constrained,NohICCV15Learning,chen_semantic_2014}. Hoffman et al.\ take each spatial unit in the fully convolutional network (FCN)~\cite{long_fully_2015} as an instance~\cite{hoffman_fcns_2016}. We contend that such instances are actually non-i.i.d.\ in either individual  domain, as their receptive fields overlap with each other. 

How can we avoid the assumption that the source and target domains share the same prediction function in a transformed domain-invariant feature space? Our proposed solution draws on two key observations. One is that the urban traffic scene images have strong idiosyncrasies (e.g., the size and spatial relations of buildings, streets, cars, etc.). Therefore, some tasks are ``easy'' and, \emph{more importantly, suffer less because of the domain discrepancy}. Second, the structured output in semantic segmentation enables convenient posterior regularization~\cite{ganchev2010posterior}, as opposed to the popular (e.g., $\ell_2$) regularization over model parameters. 

Accordingly, we propose a curriculum-style~\cite{bengio2009curriculum} domain adaptation approach. Recall that, in domain adaptation, only the source domain supplies many labeled data while there are no or only scarce labels from the target. The curriculum domain adaptation begins with the easy tasks, in order to gain some high-level properties about the unknown pixel-level labels for each target image. It then learns a semantic segmentation network --- the hard task, whose predictions over the target images are forced to follow those necessary properties as much as possible.

%is to leverage the intrinsic idiosyncrasies of the urban traffic scenes that transcend distinct domains. It is often easier to infer these properties quantitatively for the images of the target domain than to assign them pixel-wise labels. Therefore, akin to the curriculum learning~\cite{bengio2009curriculum}, we can solve the ``simple'' tasks first and then enforce the semantic segmentation models to possess the same properties on the target domain as discovered through the simple tasks. 

%To see them more clearly, we zoom out and consider the wide view. Cars drive on the road. Buildings stand beside the road. Pedestrians and cars are much smaller than buildings. Traffic lights are further smaller than people. %Such a view is typical and holds for no matter synthetic or real images.

To develop the easy tasks in the curriculum, we consider label distributions over both holistic images and some landmark superpixels of the target domain. Take the former for instance. The label distribution of an image indicates the percentage of pixels that belong to each  category, respectively. We argue that such tasks are easier, despite the domain mismatch, than assigning pixel-wise labels. Indeed, we may directly estimate the label distributions without inferring the pixel-wise labels. Moreover, the relative sizes of road, vehicle, pedestrian, etc.\ constrain the shape of the distributions, effectively reducing the search space. Finally, models to estimate the label distributions over superpixels may benefit from the urban scenes' canonical layout that transcends domains, e.g., buildings stand beside streets. 

Why and when are the seemingly simple label distributions useful for the domain adaptation of semantic segmentation? In our experiments, we find that the segmentation networks trained on the source domain perform poorly on many target images, giving rise to disproportionate label assignments (e.g., many more pixels are classified to sidewalks than to streets). To rectify this, the image-level label distribution informs the segmentation network \emph{how} to update the predictions while the label distributions of the landmark superpixels tell the network \emph{where} to update. Jointly, they guide the adaptation of the networks to the target domain to, at least, generate proportional label predictions. Note that additional ``easy tasks'' can be conveniently incorporated into our framework in the future.

%Indeed, even if the images are blurred, it is easy to tell how much proportion of the pixels corresponds to buildings. Furthermore, we argue that both humans and machine learning algorithms can give rise to more accurate predictions about such proportions by estimating them jointly than individually, because, by doing this, one naturally uses the prior about the size relations of objects in addition to the evidence of the current input image. 

%is to leverage the intrinsic idiosyncrasies of the urban traffic scenes that transcend distinct domains. It is often easier to infer these properties quantitatively for the images of the target domain than to assign them pixel-wise labels. Therefore, akin to the curriculum learning~\cite{bengio2009curriculum}, we can solve the ``simple'' tasks first and then enforce the semantic segmentation models to possess the same properties on the target domain as discovered through the simple tasks. 

Our main contribution is on  the proposed curriculum-style domain adaptation for the semantic segmentation of urban scenes. We select into the curriculum the easy and useful tasks of inferring label distributions for the target images and landmark superpixels, in order to gain some necessary properties about the target domain. Built upon these, we learn a pixel-wise discriminative segmentation network from the labeled source data and, meanwhile, conduct a ``sanity check'' to ensure the network behavior is consistent with the previously learned knowledge about the target domain. Our approach effectively eludes the assumption about the existence of a common prediction function for both domains in a transformed feature space. It readily applies to different segmentation networks, as it does not change the network architecture or tax any intermediate layers.

%Mathematically, they enable us to use the formulation in network knowledge distillation~\cite{hinton2015distilling,bucilua2006model} to conveniently incorporate the inferred target properties into the  training procedure of the semantic segmentation network. Conceptually, our domain adaptation follows a multi-stage curriculum. It first learns how to address simple tasks on the image and superpixel levels, in order to obtain  necessary properties of the target domain. Built upon those understandings, it gains pixel-wise discriminative capabilities from the labeled source data and, meanwhile, conducts ``sanity check'' to ensure the segmentation network behaves in consistency with previously learned knowledge about the target domain.

% !TEX root = main.tex

\section{Related work}
We discuss some related work on domain adaptation and semantic segmentation, with special focus on that transferring knowledge from virtual images to real photos. \vspace{-12pt}

\paragraph{Domain adaptation.} Conventional machine learning algorithms rely on the assumption that the training and test data are drawn i.i.d.\ from the same underlying distribution. However, it is often the case that there exists some discrepancy from the training to the test stage. Domain adaptation aims to rectify this mismatch and tune the models toward better generalization at testing~\cite{TorralbaCVPR11Unbiased,TommasiX15Deeper,GongLSVRR12Overcoming,KhoslaECCV12Undoing,GrettonBC09Covariate}.

%Domain adaption aims to transfer knowledge from a well practiced task to a different yet related task.

The existing work on domain adaptation mostly focuses on classification and regression problems~\cite{PatelSPM15Visual,PanTKDE10Survey}, e.g., learning from online commercial images to classify real-world objects~\cite{SaenkoECCV10Adapting,GongCVPR12Geodesic}, and, more recently, aims to improve the adaptability of deep neural networks~\cite{LongICML15Learning,GaninX15Domainadversarial,GaninICML15Unsupervised,TzengICCV15Simultaneous,bousmalis2016unsupervised,lopez2016virtual}. Among them, the most relevant work to ours is that exploring simulated data~\cite{SunBMVC14Virtual,XuX14Hierarchical,ros_synthia_2016,VazquezPAMI14Virtual,hoffman_fcns_2016,peng2017synthetic,shrivastava2016learning}. Sun and Saenko train generic object detectors from the synthetic images~\cite{SunBMVC14Virtual}, while Vazquez et al.\ use the virtual images to improve pedestrian detections in real environment~\cite{VazquezPAMI14Virtual}. The other way around, i.e., how to improve the quality of the simulated images using the real ones, is studied in~\cite{shrivastava2016learning,peng2017synthetic}. 
\vspace{-24pt}

%However, as recent data-driven models require more and more samples to train, researchers have started to address this problem with different domain adaptation approaches. Some address the problem by introducing virtual samples and minimizing virtual to real domain mismatch. 

\paragraph{Semantic segmentation.}
Semantic segmentation is the task of assigning an object label to each pixel of an image. Traditional methods~\cite{shotton_semantic_2008,tighe2010superparsing,zhang2010semantic} rely on local image features manually designed by domain experts. After the pioneering work~\cite{chen_semantic_2014,long_fully_2015} that introduced the convolutional neural network (CNN)~\cite{Lecun98Gradient} to semantic segmentation, most recent top-performing methods are built on CNNs~\cite{wu_wider_2016,ros_training_2016,badrinarayanan_segnet:_2015,zhao_pyramid_2016,NohICCV15Learning,dai_instance-aware_2016}. %due to their superior performance over traditional methods.

An enormous amount of labor-intensive work is required to annotate the many images that are needed to obtain accurate segmentation models.
The PASCAL VOC2012 Challenge~\cite{everingham_pascal_2015} contains nearly 10,000 annotated images for the segmentation competition, and the MS COCO Challenge~\cite{lin_microsoft_2014} includes over 200,000 annotated images. According to \cite{richter_playing_2016}, it took about 60 minutes to manually segment each image in \cite{brostow_semantic_2009} and about 90 minutes for each in ~\cite{cordts_cityscapes_2016}. A plausible approach to reducing the human workload is to utilize weakly supervised information such as image labels and bounding boxes~\cite{pathak2015constrained,hong_learning_2016,papandreou_weakly-and_2015,pinheiro_image-level_2015}. 

We instead explore the use of almost effortlessly labeled virtual images for training high-quality segmentation networks. In \cite{richter_playing_2016}, annotating a synthetic image took only 7 seconds on average through a computer game. For the urban scenes, we use the SYNTHIA~\cite{ros_synthia_2016} dataset which contains images of a virtual city. 
\vspace{-10pt}

% There are two sythetic dataset proposed already in 2016 \cite{ros_synthia_2016,richter_playing_2016}. A critical issue is that segmentation models trained on synthetic data are unlikely to work well on real data since there is a domain mismatch between the synthetic data and real data. And sine there will be no target annotation available under this scenarios, it would be difficult to estimate domain mismatch. 

\paragraph{Domain adaptation for semantic segmentation.} Upon observing the obvious mismatch between virtual and real data~\cite{shafaei_play_2016,richter_playing_2016,ros_synthia_2016}, we expect domain adaptation to enhance the segmentation performance on real images by networks trained on virtual ones. To the best of our knowledge, the only attempt to algorithmically address this problem is \cite{hoffman_fcns_2016}. While it regularizes the intermediate layers and constrains the output of the network, we propose a different curriculum domain adaptation strategy. We solve the easy task first and then use the learned knowledge about the target domain to regularize the network predictions.

%however, cannot use these approaches. Firstly, some have different scenarios since we have full annotation at source domain and no annotation at target domain. Secondly, they are focusing on relatively easier object semantic segmentation task where only one or two objects appear in the same image while urban scenes segmentation are full of different objects with staggered boundaries. Thirdly is that it would be difficult to have weak annotation for some large-area classes in our scenario. For example image labeled with road or bound boxes of road is simply too noisy to be transferred.

% !TEX root = main.tex

\section{Approach} \label{sApproach}

In this section, we present the details of the proposed curriculum domain adaptation for semantic segmentation of urban scene images. Unlike previous work \cite{PatelSPM15Visual,hoffman_fcns_2016} that aligns the domains via an intermediate feature space and thereby implicitly assumes the existence of the same decision function for the two domains, it is our intuition that, for structured prediction (i.e., semantic segmentation here), the cross-domain generalization of machine learning models can be more efficiently improved if we avoid this assumption and instead train them subject to necessary properties they should retain in the target domain. 
\vspace{-10pt}

\paragraph{Preliminaries.}
In particular, the properties are about the pixel-wise category labels $Y_t\in\mathbb{R}^{W\times H \times C}$ of an arbitrary image $I_t\in\mathbb{R}^{W\times H}$ from the target domain, where $W$ and $H$ are the width and height of the image, respectively, and $C$ is the number of categories. We use one-hot vector encoding for the groundtruth labels, i.e., $Y_t(i,j,c)$ takes the value of 0 or 1 and the latter means that the $c$-th label is assigned by a human annotator to the pixel at $(i,j)$. Correspondingly, the prediction $\widehat{Y}_t(i,j,c)\in[0,1]$ by a segmentation network is realized by a softmax function per pixel. 

We express each target property in the form of a distribution $p_t\in\Delta$ over the $C$ categories, where $p_t(c)$ represents the occupancy proportion of the category $c$ over the $t$-th target image or a superpixel of the image. Therefore, one can immediately calculate the distribution $p_t$ given the human annotations $Y_t$ to the image. For instance, the image level label distribution is expressed by
\begin{align}
    p_t(c) = \frac{1}{WH}\sum_{i=1}^W\sum_{j=1}^H Y_t(i,j,c), \quad \forall c. \label{eP}
\end{align}
Similarly, we can compute the target property/distribution from the network predictions $\widehat{Y}_t$ and denote it by $\widehat{p}_t$. 

\subsection{Domain adaptation using the target properties}
Ideally, we would like to have a segmentation network to imitate human annotators on the target domain. Therefore, necessarily, the properties of their annotation results should be the same too. We capture this notion by minimizing the cross entropy
$\mathcal{C}(p_t,\widehat{p}_t)=H(p_t)+\textsc{KL}(p_t,\widehat{p}_t)$ at training, where the first term of the right-hand side is the entropy and the second is the KL-divergence. 

Given a mini-batch consisting of both source images ($S$) and target images ($T$), the overall objective function for training the cross-domain generalizing segmentation network is,
\begin{align}
    \min \; \dfrac{\gamma}{|S|}\sum_{s\in S}\mathcal{L}\Big(Y_s,\widehat{Y}_s\Big) + \frac{1-\gamma}{|T|}\sum_{t\in T}\sum_{k}\mathcal{C}\Big(p^k_t,\widehat{p}^k_t\Big)  \label{eLoss}
\end{align}
where $\mathcal{L}$ is the pixel-wise cross-entropy loss defined over the sufficiently labeled source domain images, enforcing the network to have the pixel level discriminative capabilities, and the second term is over the unlabeled target domain images, hinting the network what necessary properties its predictions should have in the target domain. We use  $\gamma\in[0,1]$ to balance the two strengths in training and superscript $k$ to index different types of label distributions.

Note that in the domain adaptation context, we actually cannot directly compute the label distribution $p^k_t$ from the groundtruth annotations of the target domain. Nonetheless, estimating them using the labeled source data is easier than assigning labels to every single pixel of the target images. We present the details in the next section. 
\vspace{-10pt}

%In our experiments, we use FCN-8s~\cite{LongICCV15Fully} for the semantic segmentation network. Unlike the existing deep domain adaptation methods~\cite{GaninICML15Unsupervised,GaninX15Domainadversarial,LongICML15Learning,hoffman_fcns_2016} which introduce regularization to the intermediate layers, we only revise the loss function over the output.  Hence, our curriculum domain adaptation can be readily applied to other networks.

\paragraph{Remarks.} 
Mathematically, the objective function has a similar form as in model compression~\cite{bucilua2006model,hinton2015distilling}. We thus borrow some concepts to gain more intuitive understanding about our domain adaptation procedure. The ``student'' network follows a curriculum to learn simple knowledge about the target domain before it addresses the hard one of semantically segmenting images. The models inferring the target properties act like ``teachers'', as they hint what label distributions the final solution (image annotation) may have in the target domain at the image and superpixel levels. 

Another perspective is to understand the target properties as a posterior regularization~\cite{ganchev2010posterior} for the network. The posterior regularization can conveniently encode a priori knowledge into the objective function. Some applications include weakly supervised segmentation~\cite{pathak2015constrained,ros_training_2016} and detection~\cite{BilenBMVC14weakly}, and rule-regularized training of neural networks~\cite{hu2016harnessing}. In addition to the domain adaptation setting and novel target properties,
another key distinction of our work is that we decouple the label distributions from the network predictions and thus avoid the EM type of optimizations. Our approach learns the segmentation network with almost effortless changes to the popular deep learning tools.

\subsection{Inferring the target properties}
Thus far we have presented the ``hard'' task in the curriculum domain adaptation. In this section, we describe the ``easy'' ones, i.e., how to infer the target properties  without accessing the image annotations of the target domain. Our contributions also include selecting the particular property of label distributions to constitute the simple tasks.
\vspace{-8pt}

\subsubsection{Global label distributions of images} \label{subsLD}
Due to the domain disparity, a baseline segmentation network trained on the source domain (i.e., using the first term of eq.~(\ref{eLoss})) could be easily crippled given the target images. In our experiments, we find that our baseline network constantly mistakes streets for sidewalks and/or cars (cf.\ Figure~\ref{fQualitative}). Consequently, the predicted labels for the pixels are highly disproportionate. 

To rectify this, we employ the label distribution $p_t$ over the global image as our first property (cf.\ eq.~(\ref{eP})). Without access to the target labels, we have to train machine learning models from the labeled source images to estimate the label distribution $p_t$ for the target image. Nonetheless, we argue that this is less challenging than generating the per-pixel predictions despite that both tasks are influenced by the domain mismatch. 

%While semantic segmentation is a very computationally complex and domain sensitive problem, predicting object appearances in an image is a more relaxed and domain-invariant task given powerful image representation extracted from a well-generalized CNN such as Inception-Resnet \cite{szegedy_inception-v4_2016}. And thus it can be a useful supervision for domain adaptation.

In our experiments, we examine different  approaches to this task. We extract image features using the {Inception-Resnet-v2}~\cite{szegedy_inception-v4_2016} as the input to the following models.
\begin{description}\setlength\itemsep{0.01in}

\item[Logistic regression.] Although multinomial logistic regression (LR) is mainly used for classification, its output is actually a valid distribution over the categories. For our purpose, we thus train it by replacing the one-hot vectors in the cross-entropy loss with the groundtruth label distribution $p_s$, which is calculated using eq.~(\ref{eP}) and the available human labels of the source domain. Given a target image, we directly take the LR's output as the predicted label distribution.

\item[Mean of nearest neighbors.]
We also test a nonparametric method by simply retrieving the nearest neighbors (NNs) for a target image and then transferring the mean of the NNs' label distributions to the target image. We use the $\ell_2$ distance for the NN retrieval.

\end{description}

Finally, we include two dumb predictions as the control experiment. One is, for any target image, to output the mean of all the label distributions in the source domain (\textbf{source mean}), and the other is to output a \textbf{uniform distribution}.

%We could observe that there is a large object histogram mismatch from the table. The mismatch even enlarges from the predictions of the segmentation model is trained due to lacking of annotation and supervision in target domain. Bounding the mismatch from the neural network prediction could be a helpful weak supervision.

%We also quantitatively show the performance of our predicted histogram in Table.\ref{tLayoutmismatch}. The "NN" column is the baseline prediction method, where we assign the average histogram of $K$ (Tuned on validation dataset) nearest neighbor source images in feature space to each target image. The "LR" is the proposed  multinomial logistic regressor prediction method which yields the best performance. This shows that our histogram predictor could generate much better histogram and can thus facilitate the training.

%One may argue that this loss looks like Category Specific Adaptation term in\cite{hoffman_fcns_2016}. However their constraint peeks at the target domain annotation. Our loss also utilize each image's content to predict a unique object histogram.
\vspace{-8pt}

\subsubsection{Local label distributions of landmark superpixels} \label{subsSP}
The image level label distribution globally penalizes potentially disproportional segmentation output on the target domain, and yet is inadequate in providing spatial constraints. In this section, we consider the use of label distributions over some superpixels as the anchors to drive the network towards spatially desired target properties. 

Note that it is not necessary, and is even harmful, to use all of the superpixels in a target image to regularize the segmentation network, because that would be too strong a force and may overrule the pixel-wise discriminativeness revealed by the labeled source images, especially when the label distributions are not inferred accurately enough. 

In order to have the dual effect of both estimating the label distributions of superpixels and filtering the superpixels, we simplify the problem and employ a linear SVM in this work. In particular, we segment each image into 100 superpixels using linear spectral clustering~\cite{li_superpixel_2015}. For the superpixels of the source domain, we are able to assign a single dominant label to each of them, and then use the labels and the corresponding features extracted from the superpixels to train a multi-class SVM. Given a test superpixel of a target image, the multi-class SVM returns a class label as well as a decision value, which is interpreted as the confidence score about classifying this superpixel. We keep the top 60\% superpixels, called landmark superpixels,  in the target domain and calculate their label distributions as the second type of ``easy'' tasks. In particular, the class label of a landmark superpixel is  encoded into a one-hot vector, which serves as a valid distribution about the categories in the landmark superpixel. Albeit simple, we find this method works very well in our experiments.

We encode both visual and contextual information to represent a superpixel. First, we use the FCN-8s~\cite{long_fully_2015} pre-trained on the PASCAL CONTEXT \cite{mottaghi_role_2014} dataset, which has 59 distinct classes, to obtain 59 detection scores for each pixel. We then average them within each superpixel. Finally, we represent a superpixel by the concatenation of the 59D vectors of itself, its left and right superpixels, as well as the two respectively above and below it. 

\begin{figure}
\centering
    \includegraphics[width=0.45\textwidth]{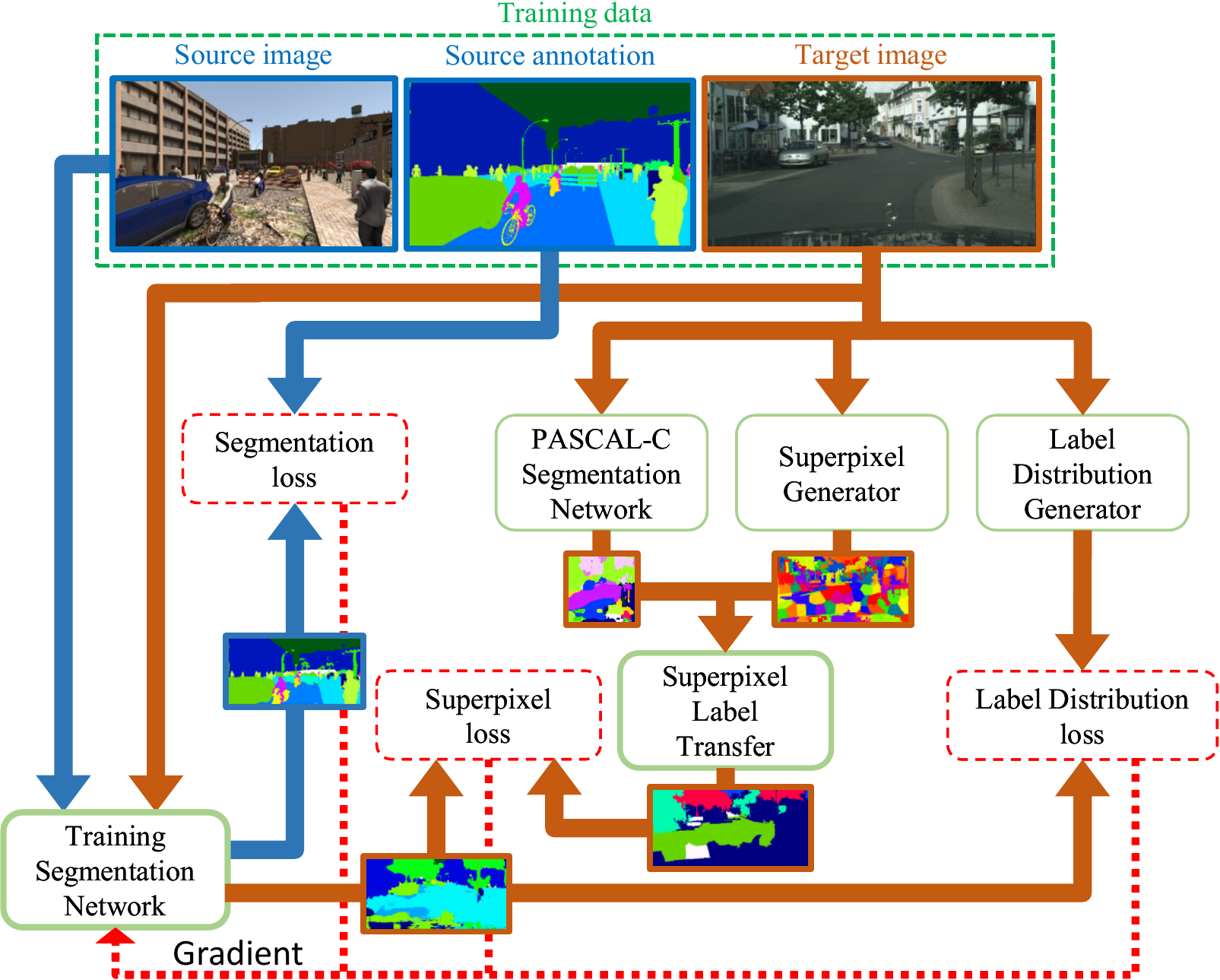}
  \caption{The overall framework of curriculum domain adaptation for semantic segmentation of urban scenes.} \label{fOverview}
\end{figure}

\subsection{Curriculum domain adaptation: recapitulation}
We recap the proposed curriculum domain adaptation using Figure~\ref{fOverview} before presenting the experiments in the next section. Our main idea is to execute the domain adaptation step by step, starting from the easy tasks that are less sensitive to the domain discrepancy than the semantic segmentation. We choose the labels distributions over global images and local landmark superpixels in this work; more tasks will be explored in the future. The solutions to them provide useful gradients originating from the target domain (cf.\ the arrows with brown color in Figure~\ref{fOverview}), while the  source domain feeds the network with well-labeled images and segmentation masks (cf.\ the dark blue arrows in Figure~\ref{fOverview}).

\begin{sidewaysfigure*}
\small
\centering
    \includegraphics[height=0.75\textheight,width=1.0\textwidth]{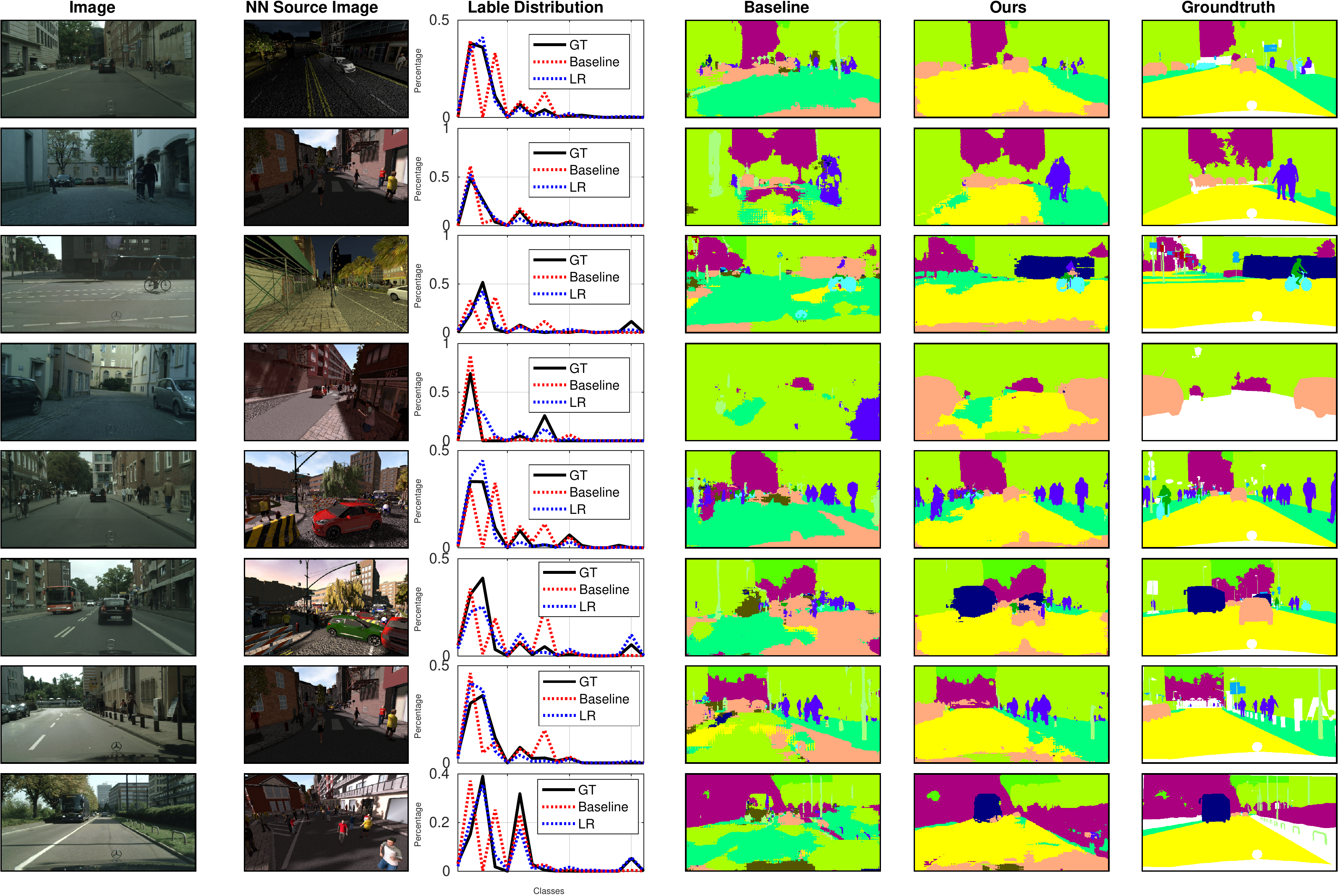}
    \vspace{-20pt}
  \caption{Qualitative semantic segmentation results on the Cityscapes dataset~\cite{ros_synthia_2016} (target domain). For each target image in the first column, we retrieve its nearest neighbor from the SYNTHIA~\cite{cordts_cityscapes_2016} dataset (source domain). The third column plots the label distributions due to the groundtruth pixel-wise semantic annotation, the predictions by the baseline network with no adaptation, and the inferred distribution by logistic regression. The last three columns are the segmentation results by the baseline network, our domain adaptation approach, and human annotators.} \label{fQualitative}
  \vspace{-15pt}
\end{sidewaysfigure*}

\section{Experiments} \label{sExperiments}

In this section, we describe the experimental setup and compare the results of our approach, its variations, and some existing baseline methods.

\subsection{Segmentation network and optimization}

In our experiments, we use FCN-8s~\cite{long_fully_2015} as our semantic segmentation network. We initialize its convolutional layers with VGG-19~\cite{SimonyanX14Very}, and then train it using the AdaDelta optimizer~\cite{zeiler_adadelta:_2012} with default parameters. Each mini-batch is comprised of five source images and five randomly chosen target images. When we train the baseline network with no adaptation, however, we try to use the largest possible mini-batch that includes 15 source images. The network is implemented in Keras~\cite{chollet_keras_2015} and Theano~\cite{al-rfou_theano:_2016}. We train different versions of the network on a single Tesla K40 GPU.

Unlike the existing deep domain adaptation methods~\cite{GaninICML15Unsupervised,GaninX15Domainadversarial,LongICML15Learning,hoffman_fcns_2016} which introduce regularization to the intermediate layers, we only revise the loss function over the output.  Hence, our curriculum domain adaptation can be readily applied to other segmentation networks (e.g.,~\cite{NohICCV15Learning,chen_semantic_2014}).

\subsection{Datasets and evaluation}
We use the publicly available \textbf{Cityscpaes}~\cite{cordts_cityscapes_2016} and \textbf{SYNTHIA}~\cite{ros_synthia_2016} datasets in our experiments. %We refer the latter as \textbf{SYNTHIA} hereon.

{Cityscapes} is a real-world, vehicle-egocentric image dataset collected in 50 cities in Germany and nearby countries. It provides four disjoint subsets: 2,993 training images, 503 validation image, 1,531 test images, and 20,021 auxiliary images. All the training, validation, and test images are accurately annotated with per pixel category labels, while the auxiliary set is coarsely labeled.  There are 34 distinct categories in the dataset.

{SYNTHIA}~\cite{ros_synthia_2016} is a large dataset of synthetic images and provides a particular subset, called SYNTHIA-RAND-CITYSCAPES, to pair with {Cityscapes}. This subset contains 9,400  images that are automatically annotated with 12 object categories,  one void class, and some unnamed classes. Note that the virtual city used to generate the synthetic images does not correspond to any of the real cities covered by {Cityscapes}. We abbreviate SYNTHIA-RAND-CITYSCAPES to {SYNTHIA} hereon.
\vspace{-10pt}

\paragraph{Domain idiosyncrasies.}
Although both datasets depict urban scenes, and {SYNTHIA} is created to be as photo-realistic as possible, they are mismatched domains in several ways. The most noticeable difference is probably the coarse-grained textures in SYNTHIA; very similar texture patterns repeat in a regular manner across different images. In contrast, the Cityscapes images are captured by high-quality dash-cameras. Another major distinction is the variability in view angles. Since Cityscapes images are recorded by the dash cameras mounted on a moving car, they are viewed from almost a constant angle that is about parallel to the ground.  More diverse view angles are employed by SYNTHIA --- it seems like some cameras are placed on the buildings that are significantly higher than a bus. Finally, some of the SYNTHIA images are severely shadowed by extreme lighting conditions, while we find no such conditions in the Cityscapes images. These combined factors, among others, make domain adaptation from SYNTHIA to Cityscapes a very challenging problem. 

Figure~\ref{fQualitative} shows some example images from both datasets. We pair each Cityscpaes image with its nearest neighbor in SYNTHIA, retrieved by the Inception-Resnet-v2 \cite{szegedy_inception-v4_2016} features. However, the cross-dataset nearest neighbors are visually very different from the query images, verifying the dramatic disparity  between the two domains.
\vspace{-10pt}

%Some prior studies on object detection~\cite{xx} and semantic segmentation~\cite{xx} on these two datasets confirm

%CityScape is a vehicle-egocentric dataset while SYNTHIA images are generated across the entire virtue city randomly. Secondly their texture is different: SYNTHIA images are synthetically generated with low resolution texture while CityScape contains high-quality dash-cam images. Thirdly their place is different: SYNTHIA dataset is generated in the virtue U.S. cities and CityScape is generated in German towns. Those combined factors makes them one of the hardest semantic segmentation domain adaption dataset intuitively and experimentally\cite{hoffman_fcns_2016}.

\paragraph{Experiment setup.}
Since our ultimate goal is to solve the semantic segmentation problem for real images of urban scenes, we take Cityscapes as the target domain and SYNTHIA as the source domain. The Cityscapes validation set is used as our test set. We split 500 images out of the Cityscpaes training set for the validation purpose (e.g., to monitor the convergence of the networks). In training,  we randomly sample mini-batches from both the images (and their labels) of SYNTHIA and the remaining images of Cityscapes yet with no labels. 

As in \cite{hoffman_fcns_2016}, we manually find 16 common classes between the two datasets: sky, building, road, sidewalk, fence, vegetation, pole, car, traffic sign, person, bicycle, motorcycle, traffic light, bus, wall, and rider. The last four are unnamed and yet labeled in SYNTHIA. 
\vspace{-10pt}

\paragraph{Evaluation.} We use the evaluation code released along with the Cityscapes dataset to evaluate our results. It calculates the PASCAL VOC intersection-over-union, i.e., 
$\text{IoU}=\frac{\text{TP}}{\text{TP}+\text{FP}+\text{FN}}$~\cite{everingham_pascal_2015}, where  TP, FP, and FN are the numbers of true positive, false positive, and false
negative pixels, respectively, determined over the whole test set. Since we have to resize the images before feeding them to the segmentation network, we resize the output segmentation mask back to the original image size before running the evaluation against the groundtruth annotations.

%Besides class accuracy, we evaluate our model over the widely-used mean intersection-over-union (meanIU) metric proposed in PASCAL semantic segmentation challenge~\cite{EveringhamIJCV10Pascal}. The meanIU score is generated by the official CityScape evaluation toolbox on our validation prediction and groundtruth. We resize our validation prediction to original Cityscape image resolution before evaluation to ensure the most accurate results.

\begin{table}
\centering
\caption{The $\chi^2$ distances between the groundtruth label distributions and those predicted by different methods.}
\label{tLayoutmismatch}
\scalebox{0.9}{
\begin{tabular}{l|ccccc}
\hline
Method   & Uniform & NoAdapt & Src mean  & NN & \textbf{LR}\\
\hline
$\chi^2$ Distance   & 1.13 & 0.65 & 0.44  & 0.33 & \textbf{0.27}\\
\hline
\end{tabular}
}
\vspace{-15pt}
\end{table}

\subsection{Results of inferring global label distributions}
Before presenting the final semantic segmentation results, we first compare the different approaches to inferring the global label distributions of the target images (cf.\ Section~\ref{subsLD}). We report the results on the held-out validation images of Cityscapes in this experiment, and then select the best method for the remaining experiments. 

In Table~\ref{tLayoutmismatch}, we compare the estimated label distributions with the groundtruth ones using the $\chi^2$ distance, the smaller the better. We see that the baseline network (NoAdapt), which is directly learned from the source domain without any adaptation methods, outperforms the dumb uniform distribution (Uniform) and yet no other methods. This confirms that the baseline network gives rise to severely disproportionate predictions over the target domain. 

Another dumb prediction (Src mean), i.e., using the mean of  all label distributions over the source domain as the prediction for the target images, however, performs reasonably well. To some extent, this indicates the value of the simulated source domain for the semantic segmentation task of urban scenes. 

Finally, the nearest neighbors (NN) based method and the multinomial logistic regression (LR) (cf.\ Section~\ref{subsLD}) perform the best. We use the output of LR on the target domain in our remaining experiments.

\subsection{Comparison results}

\begin{table*}
    \centering
    \small
\caption{Comparison results for the semantic segmentation of the Cityscapes images~\cite{cordts_cityscapes_2016} by adapting from SYNTHIA~\cite{ros_synthia_2016}.}
\vspace{-3pt}
\label{Tresults}
\scalebox{0.92}{
\begin{tabular}{l||c||c|c|c|c|c|c|c|c|c|c|c|c|c|c|c|c}
\hline
\multirow{2}{*}[-2em]{ Method~~~~\%} & \multirow{2}{*}[-2em]{ IoU} & \multicolumn{16}{c}{Class-wise IoU}\\

 & &  \rot{bike} & \rot{fence} & \rot{wall} & \rot{t-sign} & \rot{pole} & \rot{mbike} & \rot{t-light} & \rot{sky} & \rot{bus} & \rot{rider} & \rot{veg} & \rot{bldg} & \rot{car} & \rot{person} & \rot{sidewalk} & \rot{road}\\
 \hline\hline
 NoAdapt~\cite{hoffman_fcns_2016}  & 17.4 & 0.0 & 0.0 & 1.2 & 7.2 & 15.1 & 0.1 & 0.0 & 66.8 & \underline{\textit{3.9}} & 1.5 & 30.3 & 29.7 & 47.3 & 51.1 & 17.7 & 6.4\\

FCN Wld~\cite{hoffman_fcns_2016} & \underline{\textit{20.2}} & \underline{\textit{0.6}} & \underline{\textit{0.0}} & \textbf{4.4} & \textbf{11.7} & \underline{\textit{20.3}} & \underline{\textit{0.2}} & \underline{\textit{0.1}} & \underline{\textit{68.7}} & 3.2 & \underline{\textit{3.8}} & \underline{\textit{42.3}} & \underline{\textit{30.8}} & \textbf{54.0} & \textbf{51.2} & \underline{\textit{19.6}} & \underline{\textit{11.5}}\\
\hline
NoAdapt & 22.0 & \textbf{18.0} & 0.5 & 0.8 & 5.3 & 21.5 & 0.5 & 8.0 & \underline{\textit{75.6}} & 4.5 & \textbf{9.0} & \underline{\textit{72.4}} & 59.6 & 23.6 & \underline{\textit{35.1}} & 11.2 & 5.6\\

\textbf{Ours (I)} & \underline{\textit{25.5}} & 16.7 & \textbf{0.8} & \underline{\textit{2.3}} & \underline{\textit{6.4}} & \textbf{21.7} & \textbf{1.0} & \textbf{9.9} & 59.6 & \underline{\textit{12.1}} & 7.9 & 70.2 & \underline{\textit{67.5}} & \underline{\textit{32.0}} & 29.3 & \underline{\textit{18.1}} & \underline{\textit{51.9}}\\
\hline
SP Lndmk & { 23.0} & { 0.0} & { 0.0} & { 0.0} & { 0.0} & { 0.0} & { 0.0} & { 0.0} & \textbf{ 83.1} & \textbf{ 26.1} & { 0.0} & { 73.1} & { 67.7} & { 41.1} & { 5.8} & { 10.6} & { 60.8}\\

SP & { 25.6} & { 0.0} & { 0.0} & { 0.0} & { 0.0} & { 0.0} & { 0.0} & { 0.0} & { 80.5} & { 22.1} & { 0.0} & { 71.9} & { 69.3} &  \underline{\textit{45.9}} & { 24.6} & { 19.8} & \textbf{ 75.0}\\

\textbf{Ours (SP)} & \underline{\textit{28.1}} & \underline{\textit{10.2}} & \underline{\textit{0.4}} & \underline{\textit{0.1}} & \underline{\textit{2.7}} & \underline{\textit{8.1}} & \underline{\textit{0.8}} & \underline{\textit{3.7}} & 68.7 & 21.4 & \underline{\textit{7.9}} & \underline{\textit{75.5}} & \underline{\textit{74.6}} & 42.9 & \underline{\textit{47.3}} & \underline{\textit{23.9}} & 61.8\\
\hline
\textbf{Ours (I+SP)} & \textbf{29.0} & 13.1 & 0.5 & 0.1 & 3.0 & 10.7 & 0.7 & 3.7 & 70.6 & 20.7 & 8.2 & \textbf{76.1} & \textbf{74.9} & 43.2 & 47.1 & \textbf{26.1} & 65.2\\
\hline
\end{tabular}
}
\vspace{-8pt}
\end{table*}

We report the final semantic segmentation results on the test data of the target domain in this section. We compare our approach to the following competing methods.

\begin{description}
  \setlength\itemsep{0.01in}

\item[No adaptation (NoAdapt).] We directly train the FCN-8s model on SYNTHIA without applying any domain adaptation methods. This is the most basic baseline for our experiments. 

\item[Superpixel classification (SP).]\label{baselineSP} Recall that we have trained a multi-class SVM using the dominant labels of the superpixels in the source domain. We then use them to classify the target superpixels. 

\item[Landmark superpixels (SP Lndmk).] Since we keep the top 60\% most confidently classified superpixels  as the landmarks to regularize our segmentation network during training (cf.\ Section~\ref{subsSP}), it is also interesting to examine the classification results of these superpixels. We run the evaluation after assigning the void class label to the other pixels of the images. 

In addition to the IoU, we have also evaluated the classification results of the superpixels by accuracy. We find that the classification accuracy is 71\% for all the superpixels of the target domain, while for the selected 60\% landmark superpixels, the classification accuracy is more than 88\%. 

%\textbf{Refined SP} is the 30\%  most confident superpixels subsampled from Completed SP and used in the network training. Note that unlike Completed SP, refined SP does not cover the entire image. Hence it will has lower performance in semantic segmentation. However the quality of the superpixel annotation is higher. The accuracy of Completed SP is 71\% and the accuracy of Refined SP is 88\%.

\item[FCNs in the wild (FCN Wld).] Hoffman et al.'s work~\textbf{\cite{hoffman_fcns_2016}} is the only existing one addressing the same problem as ours, to the best of our knowledge. They introduce a pixel-level adversarial loss to the intermediate layers of the network and impose constraints to the network output. Their experimental setup is about identical to ours except that they do not specify which part of Cityscapes is considered as the test set. Nonetheless, we include their results for comparison to put our work in a better perspective. 
%dataset they are running their experiment on. We directly put their numbers onto our figure after removing the redundant classes recalculating the meanIU.

\end{description}

The comparison results are shown in Table~\ref{Tresults}. Immediately, we note that all our domain adaptation results are significantly better than those without adaptation (NoAdapt). 

We denote by (\textbf{Ours (I)}) the network trained using the global label distributions over the target images (and the labeled source images). Although one may wonder that the image-wise label distributions are too high-level to supervise the pixel-wise discriminative network, the gain is actually significant. They are able to correct some obvious errors of the baseline network, such as the disproportional predictions about road and sidewalk (cf.\ the results of \textbf{Ours (I)} vs.\ NoAdapt in the last two columns).

It is interesting to see that both superpixel classification-based segmentation results (SP and SP Lndmk) are also better than the baseline network (NoAdapt). The label distributions obtained over the landmark superpixels boost the segmentation network (\textbf{Ours (SP)}) to the mean IoU of 28.1\%, which is better than those by either superpixel classification or the baseline network individually. We have also tried to use the label distributions over all the superpixels to train the network, and observe little improvement over NoAdapt. This is probably because it is too forceful to regularize the network output at every single superpixel especially when the estimated label distributions are not accurate enough.

The superpixel-based methods, including \textbf{Ours (SP)}, miss small objects such as fences, traffic lights (t-light), and traffic signs (t-sign), and instead are very accurate for categories like the sky, road, and building, that typically occupy larger image regions. On the contrary, the label distributions on the images give rise to a network (\textbf{Ours (I)}) that performs better on the small objects than \textbf{Ours (SP)}. In other words, they mutually complement to some extent. Re-training the network by using the label distributions over both global images and local landmark superpixels (\textbf{Ours (I+SP)}), we achieve the best semantic segmentation results on the target domain. In the future work, it is worth exploring other target properties, perhaps still in the form of label distributions, that handle the small objects well, in order to further complement the superpixel-level label distributions.
\vspace{-15pt}

\paragraph{Comparison with FCNs in the wild~\cite{hoffman_fcns_2016}.} 
Although we use the same segmentation network (FCN-8s) as~\cite{hoffman_fcns_2016},  our baseline results (NoAdapt) are better than those reported in~\cite{hoffman_fcns_2016}. This may be due to subtle differences in terms of implementation or experimental setup. Although our own baseline results are superior,  we gain larger improvements (7\%) over them than the performance gain of~\cite{hoffman_fcns_2016} (3\%) over the seemingly underperforming baseline network there. 
\vspace{-10pt}

\paragraph{Comparison with learning domain-invariant features.} At our first attempt to solve the domain adaptation problem for the semantic segmentation of urban scenes, we tried to learn domain invariant features following the deep domain adaptation methods~\cite{LongICML15Learning} for classification. In particular, we impose the maximum mean discrepancy~\cite{gretton2012kernel}  over the layer before the output. We name such network layer the feature layer. Since there are virtually three output layers in FCN-8s, we experiment with all the three feature layers correspondingly. We have also tested the domain adaptation by reversing the gradients of a domain classifier~\cite{GaninICML15Unsupervised}. However, none of these efforts lead to any noticeable gain over the baseline network so the results are omitted.

%We could also observe that our superpixel adapted network has better performance than the completed superpixels themselves. This indicates that although the superpixel label transfer technique could serve as a powerful baseline, it is not as powerful as our method only trained on superpixel loss. It is also obvious that our network outperformed the refined superpixel baseline, which provides the annotation for the superpixel adapted network. This empirically validates that the task of refined superpixels is to bridging the appearance information between the source domain and target domain rather than providing direct supervision.

%constraint improve the segmentation performance on small objects compared with network trained with superpixel constraint and baseline. It also encourages under-predicted classes. In the baseline, the network rarely classify any pixels into road and resulted in low performance in road. However introducing layout constraint greatly improved the performance on road class by 40\%.

%Finally, our combined results achieved the best mean performance.

%completed superpixels and refined ones does not annotate small object classes such as traffic light and pole etc. This is reasonable because the superpixels can hardly segments those small and shape objects correctly. This results in lacking supervision in the superpixel adapted network. However the superpixel constraint greatly boost the segmentation performance on the large objects. 

% !TEX root = main.tex

\section{Conclusion}

In this paper, we address domain adaptation for the semantic segmentation of urban scenes. We propose a curriculum style approach to this problem. We learn to estimate the global label distributions of the images and local label distributions of the landmark superpixels of the target domain. Such tasks are easier to solve than the pixel-wise label assignment. Therefore, we  use their results to effectively regularize our training of the semantic segmentation network such that its predictions meet the inferred label distributions over the target domain. Our method outperforms several competing methods that do domain adaptation from simulated images to real photos of urban traffic scenes. In future work, we will explore more target properties that can be conveniently inferred to enrich our curriculum domain adaptation framework.

\vspace{-10pt}
\paragraph{Acknowledgements.}
This work is supported by the NSF award IIS \#1566511, a gift from Adobe Systems Inc., and a GPU from NVIDIA. We thank the anonymous reviewers and area chairs for their insightful comments.

%We propose to constrain the CNN training process with global label distribution and local landmark superpixel. Our experiment results indicates that our network significantly outperform both the non-domain-adapted baseline and the only one addressing this problem to our best knowledge. Our insight on landmark superpixel and label distribution could help this emerging research direction.

{\small
\bibliographystyle{ieee}
\bibliography{egbib}
}

% !TEX root = main.tex

\appendix

\begin{table*}
    \centering
    \small
\caption{Comparison results for the semantic segmentation of the Cityscapes images~\cite{cordts_cityscapes_2016} by adapting from GTA~\cite{richter_playing_2016}.}
\label{Tresults_GTA}
\scalebox{0.85}{
\begin{tabular}{l||c||c|c|c|c|c|c|c|c|c|c|c|c|c|c|c|c|c|c|c}

\hline
\multirow{2}{*}[-2em]{ Method~~~~\%} & \multirow{2}{*}[-2em]{ IoU} & \multicolumn{19}{c}{ Class-wise IoU}\\

 & &  \rot{bike} & \rot{fence} & \rot{wall} & \rot{t-sign} & \rot{pole} & \rot{mbike} & \rot{t-light} & \rot{sky} & \rot{bus} & \rot{rider} & \rot{veg}  & \rot{terrain} & \rot{train} & \rot{bldg} & \rot{car} & \rot{person} & \rot{truck} & \rot{sidewalk} & \rot{road}\\
 \hline\hline
 NoAdapt~\cite{hoffman_fcns_2016}  & 21.1 & 0.0 & 3.1 & 7.4 & 1.0 & 16.0 & 0.0 & 10.4 & 58.9 & 3.7 & 1.0 & 76.5 & 13 & 0.0 & 47.7 & 67.1 & 36 & 9.5 & 18.9 & 31.9\\

FCN Wld~\cite{hoffman_fcns_2016} & 27.1 & 0.0 & 5.4 & \textbf{14.9} & 2.7 & 10.9 & 3.5 & 14.2 & 64.6 & 7.3 & 4.2 & \textbf{79.2} & 21.3 & 0.0 & 62.1 & \textbf{70.4} & \textbf{44.1} & 8.0 & \textbf{32.4} & 70.4\\
\hline
NoAdapt & 22.3 & 13.8 & 8.7 & 7.3 & \textbf{16.8} & \textbf{21.0} & 4.3 & 14.9 & 64.4 & 5.0 & \textbf{17.5} & 45.9 & 2.4 & 6.9 & 64.1 & 55.3 & 41.6 & 8.4 & 6.8 & 18.1\\
\hline
\textbf{Ours (I)} & 23.1 & 9.5 & 9.4 & 10.2 & 14.0 & 20.2 & 3.8 & 13.6 & 63.8 & 3.4 & 10.6 & 56.9 & 2.8 & \textbf{10.9} & 69.7 & 60.5 & 31.8 & 10.9 & 10.8 & 26.4\\
\hline
\textbf{Ours (SP)} & 27.8 & \textbf{15.6} & 11.7 & 5.7 & 12.0 & 9.2 & 12.9 & 15.5 & 64.9 & 15.5 & 9.1 & 74.6 & 11.1 & 0.0 & 70.5 & 56.1 & 34.8 & 15.9 & 21.8 & 72.1\\
SP Lndmk & 21.4 & 0.0 & 0.0 & 0.0 & 0.0 & 0.0 & 0.0 & 0.0 & \textbf{82.9} & 10.0 & 0.0 & 74.5 & 22.5 & 0.0 & 69.9 & 52.7 & 13.1 & 11.2 & 8.0 & 61.8\\

SP & 26.8 & 0.3 & 4.1 & 7.6 & 0.0 & 0.2 & 0.9 & 0.0 & 81.6 & \textbf{25.3} & 3.5 & 73.0 & \textbf{32.1} & 0.0 & 71.0 & 61.9 & 26.2 & \textbf{30.4} & 19.2 & 71.8 \\
\hline
\textbf{Ours (I+SP)} & \textbf{28.9} & 14.6 & \textbf{11.9} & 6.0 & 11.1 & 8.4 & \textbf{16.8} & \textbf{16.3} & 66.5 & 18.9 & 9.3 & 75.7 & 13.3 & 0.0 & \textbf{71.7} & 55.2 & 38.0 & 18.8 & 22.0 & \textbf{74.9}\\

\hline
\end{tabular}
}
\end{table*}

\eat{
\begin{table*}[t!]
    \centering
    \small
\caption{Ablation study about the number of superpixels on SYNTHIA$\rightarrow$Cityscapes.}
\label{Tresults_SPnum}
\scalebox{0.85}{
\begin{tabular}{l||c||c|c|c|c|c|c|c|c|c|c|c|c|c|c|c|c}

\hline
\multirow{2}{*}[-2em]{ SP Number~~~~\%} & \multirow{2}{*}[-2em]{ IoU} & \multicolumn{16}{c}{Class-wise IoU}\\

 & &  \rot{bike} & \rot{fence} & \rot{wall} & \rot{t-sign} & \rot{pole} & \rot{mbike} & \rot{t-light} & \rot{sky} & \rot{bus} & \rot{rider} & \rot{veg} & \rot{bldg} & \rot{car} & \rot{person} & \rot{sidewalk} & \rot{road}\\
 \hline\hline
50 & 25.1 & 0.0 & 0.0 & 0.0 & 0.0 & 0.0 & 0.0 & 0.0 & 82.3 & 24.5 & 0.0 & 69.5 & 66.7 & 44.9 & 21.5 & 18.2 & 74.2\\
100 & { 25.6} & { 0.0} & { 0.0} & { 0.0} & { 0.0} & { 0.0} & { 0.0} & { 0.0} & { 80.5} & { 22.1} & { 0.0} & { 71.9} & { 69.3} & 45.9 & { 24.6} & { 19.8} & 75.0\\
200 & 26.5 & 0.0 & 0.0 & 0.0 & 0.0 & 0.0 & 0.0 & 0.0 & 83.2 & 25.9 & 0.0 & 75.5 & 68.8 & 48.7 & 29.7 & 22.2 & 70.4 \\

\hline
\end{tabular}
}
\end{table*}
}

\section*{GTA$\rightarrow$Cityscapes}

The main text above has been accepted to IEEE International Conference on Computer Vision (ICCV) 2017. After the paper submission, we have been continuously working on the project and have got more results. We include them below to complement the experiments in  the main text.

The new experiment is basically the same as the one in the main text except that we replace SYNTHIA with the GTA dataset~\cite{richter_playing_2016}. {GTA} is a synthetic, vehicle-egocentric image dataset collected from the open world in the realistically rendered computer game Grand Theft Auto V (GTA, or GTA5). It contains 24,996 images, whose semantic segmentation annotations are fully compatible with the classes used in Cityscapes. Hence we use all the 19 official training classes in our experiment. The results are shown in Table \ref{Tresults_GTA}.

As in the main text, the same observations about our approach apply here. Additionally, we note that the results are overall better than those adapting from SYNTHIA to Cityscapes. This is not surprising, because the GTA images are more photo-realistic than SYNTHIA's. 

%This is not only backed by the visual observation but also segmentation results shown in the table. Since SYNTHIA only has 16 classes while GTA has 19 classes, segmenting CityScape image using GTA trained model would be harder due to three extra classes. Despite this, its mean IoU is still equivalent to SYNTHIA2CityScape experiment. And without surprise, the \textbf{Ours (I+SP)} outperform all other baselines in mean IoU. Unlike SYNTHIA2CityScape scenario, \textbf{Ours (I+SP)} has the best performance in detecting some tiny classes such as fence, motor bike and traffic light.

%\section{Superpixel Granularity}

%As superpixel shrink and become more fine-grained, they could represent object boundary and small objects more precisely. How would different superpixels impact the performance?

%To quantitatively look into this problem, we predict CityScape segmentation with superpixel classifier trained on SYNTHIA images. In this experiment we vary the number of training \ testing superpixels in source \ target domain. As shown in Table \ref{Tresults_SPnum}, although our default superpixels is 100 per image, increase the superpixel could definitely increase the the accuracy of superpixel prediction without allocating more resource for training neural network.

\end{document}

\newcommand*\rot{\rotatebox{90}}

\newcommand{\BG}[1]{\textcolor{blue}{BG: #1}}
\newcommand{\PD}[1]{\textcolor{blue}{PD: #1}}
\newcommand{\YZ}[1]{\textcolor{blue}{YZ: #1}}

%%%%%%%%% TITLE
\title{Supplement Materials for\\ Curriculum Domain Adaptation for Semantic Segmentation of Urban Scenes}

\author{First Author\\
Institution1\\
Institution1 address\\
{\tt\small firstauthor@i1.org}
\and
Second Author\\
Institution2\\
First line of institution2 address\\
{\tt\small secondauthor@i2.org}
}

\maketitle
%\thispagestyle{empty}

%\section{Additional Experiment}

We consider a new source domain in the supplementary material, which is also comprised of simulated images of inner-city street scenes.  The experiment shares the same evaluation metric, target dataset, segmentation network, optimization protocol, testing and validation dataset, and baseline models as the one in our main paper. Other details also remain the same  unless otherwise specified. %The only differences are the object classes and source dataset. 

\section{Datasets and classes}

The target domain is still the \textbf{Cityscpaes}~\cite{cordts_cityscapes_2016} dataset as in our main paper. However we switch our source domain to the \textbf{GTA} dataset~\cite{richter_playing_2016}.  {GTA} is a synthetic, vehicle-egocentric image dataset collected from the realistically rendered computer game Grand Theft Auto (GTA). It contains 24,996 images and each is annotated with pixel-wise semantic labels.

\paragraph{Experiment setup.}

Since the annotation in our source dataset - GTA dataset is fully compatible with our target dataset - CityScape dataset, we will train and evaluate our model on the default 19 officially-suggested CityScape training classes:  sky, building, road, sidewalk, fence, vegetation, pole, car, traffic sign, person, bicycle, motorcycle, traffic light, bus, wall, rider, train, truck and terrain.

\paragraph{Domain idiosyncrasies.}

Although both SYNTHIA~\cite{ros_synthia_2016} and GTA dataset are synthetic data, the GTA dataset is more visually similar to CityScape due to the following reasons. First of all, textures and shadows in GTA dataset are much more realistically rendered thanks to the state-of-the-art commercial game engine. Secondly, just like the CityScape dataset, the GTA dataset images are also taken by a dash camera on a ``car'' traveling in the virtual city of GTA game. In this case, although the domain adaptation task is still challenging, the aforementioned factors make it a presumably easier task to adapt from GTA to Cityscapes than from SYNTHIA to Cityscapes.

\subsection{Comparison results}

\begin{table*}
    \centering
    \small
\caption{Comparison results for the semantic segmentation of the Cityscapes images~\cite{cordts_cityscapes_2016}.}
\label{Tresults}
\scalebox{0.85}{
\begin{tabular}{l||c||c|c|c|c|c|c|c|c|c|c|c|c|c|c|c|c|c|c|c}

\hline
\multirow{2}{*}[-2em]{ Method~~~~\%} & \multirow{2}{*}[-2em]{ IoU} & \multicolumn{19}{c}{Class-wise IoU}\\

 & &  \rot{bike} & \rot{fence} & \rot{wall} & \rot{t-sign} & \rot{pole} & \rot{mbike} & \rot{t-light} & \rot{sky} & \rot{bus} & \rot{rider} & \rot{veg}  & \rot{terrain} & \rot{train} & \rot{bldg} & \rot{car} & \rot{person} & \rot{truck} & \rot{sidewalk} & \rot{road}\\
 \hline\hline
 NoAdapt~\cite{hoffman_fcns_2016}  & 21.1 & 0.0 & 3.1 & 7.4 & 1.0 & 16.0 & 0.0 & 10.4 & 58.9 & 3.7 & 1.0 & 76.5 & 13 & 0.0 & 47.7 & 67.1 & 36 & 9.5 & 18.9 & 31.9\\

FCN Wld~\cite{hoffman_fcns_2016} & 27.1 & 0.0 & 5.4 & \textbf{14.9} & 2.7 & 10.9 & 3.5 & 14.2 & 64.6 & 7.3 & 4.2 & \textbf{79.2} & 21.3 & 0.0 & 62.1 & \textbf{70.4} & \textbf{44.1} & 8.0 & \textbf{32.4} & 70.4\\
\hline
NoAdapt & 22.3 & 13.8 & 8.7 & 7.3 & \textbf{16.8} & \textbf{21.0} & 4.3 & 14.9 & 64.4 & 5.0 & \textbf{17.5} & 45.9 & 2.4 & 6.9 & 64.1 & 55.3 & 41.6 & 8.4 & 6.8 & 18.1\\

\textbf{Ours (CC)} & 23.1 & 9.5 & 9.4 & 10.2 & 14.0 & 20.2 & 3.8 & 13.6 & 63.8 & 3.4 & 10.6 & 56.9 & 2.8 & \textbf{10.9} & 69.7 & 60.5 & 31.8 & 10.9 & 10.8 & 26.4\\

\textbf{Ours (I)} & 23.1 & 9.5 & 9.4 & 10.2 & 14.0 & 20.2 & 3.8 & 13.6 & 63.8 & 3.4 & 10.6 & 56.9 & 2.8 & \textbf{10.9} & 69.7 & 60.5 & 31.8 & 10.9 & 10.8 & 26.4\\
\hline
SP Lndmk & 19.1 & 0.0 & 0.0 & 0.0 & 0.0 & 0.0 & 0.0 & 0.0 & \textbf{82.6} & 0.3 & 0.0 & 73.6 & 17.1 & 0.0 & 65.1 & 47.2 & 7.9 & 10.3 & 3.5 & 54.8\\

SP & 26.8 & 0.3 & 4.1 & 7.6 & 0.0 & 0.2 & 0.9 & 0.0 & 81.6 & \textbf{25.3} & 3.5 & 73.0 & \textbf{32.1} & 0.0 & 71.0 & 61.9 & 26.2 & \textbf{30.4} & 19.2 & 71.8\\

\textbf{Ours (SP)} & 27.8 & \textbf{15.6} & 11.7 & 5.7 & 12.0 & 9.2 & 12.9 & 15.5 & 64.9 & 15.5 & 9.1 & 74.6 & 11.1 & 0.0 & 70.5 & 56.1 & 34.8 & 15.9 & 21.8 & 72.1\\
\hline
\textbf{Ours (I+SP)} & \textbf{28.9} & 14.6 & \textbf{11.9} & 6.0 & 11.1 & 8.4 & \textbf{16.8} & \textbf{16.3} & 66.5 & 18.9 & 9.3 & 75.7 & 13.3 & 0.0 & \textbf{71.7} & 55.2 & 38.0 & 18.8 & 22.0 & \textbf{74.9}\\

\hline
\end{tabular}
}
\vspace{-8pt}
\end{table*}

We present our results in Table~\ref{Tresults}.

{\small
\bibliographystyle{ieee}
\bibliography{egbib}
}